\documentclass[twoside]{article}

\usepackage[accepted]{aistats2018}
\usepackage{amsmath}
\usepackage{amssymb}
\usepackage{graphicx}
\usepackage{subfigure}
\usepackage{algorithm}
\usepackage{algorithmic}
\usepackage{widetext}
\usepackage{cuted}
\usepackage{bm}

\DeclareMathOperator{\htanh}{htanh}
\DeclareMathOperator{\tr}{tr}
\DeclareMathOperator{\erf}{erf}
\DeclareMathOperator{\erfc}{erfc}

\newcommand{\A}{\mathbf{A}}
\newcommand{\mb}[1]{\mathbf{#1}}

\begin{document}

\twocolumn[

\aistatstitle{The Emergence of Spectral Universality in Deep Networks}

\aistatsauthor{ Jeffrey Pennington \And Samuel S. Schoenholz \And  Surya Ganguli }

\aistatsaddress{ Google Brain \And  Google Brain \And Google Brain\\Applied Physics, Stanford University }  ]

\begin{abstract}
Recent work has shown that tight concentration of the {\it entire} spectrum of singular values of a deep network's input-output Jacobian around one at initialization can speed up learning by orders of magnitude. Therefore, to guide important design choices, it is important to build a full theoretical understanding of the spectra of Jacobians at initialization. To this end, we leverage powerful tools from free probability theory to provide a detailed analytic understanding of how a deep network's Jacobian spectrum depends on various hyperparameters including the nonlinearity, the weight and bias distributions, and the depth. For a variety of nonlinearities, our work reveals the emergence of new universal limiting spectral distributions that remain concentrated around one even as the depth goes to infinity.
\end{abstract} 

\section{INTRODUCTION}

A well-conditioned initialization is essential for successfully training neural networks.  Seminal initial work focused on random weight initializations ensuring that the second moment of the spectrum of singular values of the network Jacobian from input to output remained one, thereby preventing exponential explosion or vanishing of gradients \cite{glorot2010understanding}.  However, recent work has shown that even among different random initializations sharing this property,  those whose {\it entire} spectrum tightly concentrates around one can often yield faster learning by orders of magnitude.  For example, deep linear networks with orthogonal initializations, for which the entire spectrum is exactly one, can achieve depth-independent learning speeds, while the corresponding Gaussian initializations cannot~\cite{saxe2013}. 

Recently, it was shown~\cite{pennington2017resurrecting} that a similarly well-conditioned Jacobian could be constructed for deep non-linear networks using a combination of orthogonal weights and $\tanh$ nonlinearities. The result of this improved conditioning was an orders-of-magnitude speedup in learning for $\tanh$ networks. However, the same study also proved that a well-conditioned Jacobian could not be achieved with Rectified Linear units (ReLUs). Together these results explained why, historically, in some cases orthogonal weight initialization had been found to improve training efficiency only slightly~\cite{mishkin2015}.

These empirical results connecting the conditioning of the Jacobian to a dramatic speedup in learning raise an important theoretical question. Namely, how does the entire shape of this spectrum depend on a network's nonlinearity, weight and bias distribution, and depth?  Here we provide a detailed analytic answer by using powerful tools from free probability theory.  Our answer provides theoretical guidance on how to choose these different network ingredients so as to achieve tight concentration of deep Jacobian spectra even at very large depths.  Along the way, we find several surprises, and we summarize our results in the discussion.      

\section{PRELIMINARIES}

\subsection{Problem Setup}

Consider an $L$-layer feed-forward neural network of width $N$ with synaptic weight matrices $\mb{W}^l\in \mathbb{R}^{N\times N}$,  bias vectors $\mb{b}^l$, pre-activations $\mb{h}^l$, and post-activations $\mb{x}^l$, with $l=1,\dots,L$.  The forward-propagation dynamics are given by,
\begin{equation}
\label{eqn:dynam}
\mb{x}^l = \phi(\mb{h}^l)\,,\quad \mb{h}^l = \mb{W}^l \mb{x}^{l-1} + \mb{b}^l\,,
\end{equation}
where $\phi : \mathbb{R} \to \mathbb{R}$ is a pointwise nonlinearity and the input is  $\mb{x}^0 \in \mathbb{R}^N$.  Now consider the input-output Jacobian $\mb{J} \in \mathbb{R}^{N\times N}$ given by
\begin{equation}
\begin{split}
\label{eqn:Jz}
\mb{J} = \frac{\partial \mb{x}^L}{\partial \mb{x}^0} = \prod_{l=1}^L \mb{D}^l \mb{W}^l.
\end{split}
\end{equation}
Here $\mb{D}^l$ is a diagonal matrix with entries $D^l_{ij} = \phi'(h^l_i) \, \delta_{ij}$, where $\delta_{ij}$ is the Kronecker delta function.
The input-output Jacobian $\mb{J}$ is closely related to the backpropagation operator mapping output errors to weight matrices at a given layer, in the sense that if the former is well-conditioned, then the latter tends to be well-conditioned for all weight layers. We are therefore interested in understanding the entire singular value spectrum of $\mb{J}$ for deep networks with randomly initialized weights and biases.

In particular, we will take the biases $\mb{b}^l_i$ to be drawn i.i.d. from a zero-mean Gaussian with standard deviation $\sigma_b$.  For the weights, we will consider two random matrix ensembles: (1) random {\it Gaussian} weights in which each $W^l_{ij}$ is drawn i.i.d from a Gaussian with variance $\sigma_w^2/N$, and (2) random {\it orthogonal} weights, drawn from a uniform distribution over scaled orthogonal matrices obeying  $(\mb{W}^{l})^T \mb{W}^l = \sigma_w^2 \, \mb{I}$.

\subsection{Review of Signal Propagation}

The random matrices $\mb{D}^l$ in \eqref{eqn:Jz} depend on the empirical distribution of pre-activations $h^l_i$ for $i=1,\dots,N$ entering the nonlinearity $\phi$ in \eqref{eqn:dynam}.  The propagation of this empirical distribution through different layers $l$ was studied in \cite{poole2016, schoenholz2016}. In those works, it was shown that in the large $N$ limit this empirical distribution converges to a Gaussian with zero mean and variance $q^l$, where $q^l$ obeys a recursion relation induced by the dynamics in \eqref{eqn:dynam}:
\begin{equation}
q^l \, = \,  \sigma_w^2 \int \, \mathcal{D}h  \, \phi \left( \sqrt{q^{l-1}} h \right)^2  + \sigma_b^2\,, 
\end{equation}
with initial condition $q^1 = \frac{\sigma_w^2}{N} \sum_{i=1}^N (x^0_i)^2+\sigma_b^2$,
and $\mathcal{D}h = \frac{dh}{\sqrt{2\pi}} \,\exp{(-\frac{h^2}{2})}$ denoting the standard normal measure. This recursion has a fixed point obeying, 
\begin{equation}
q^* \, = \,  \sigma_w^2 \int \, \mathcal{D}h  \, \phi \left( \sqrt{q^*} h \right)^2  + \sigma_b^2\,.
\label{eq:qliter}
\end{equation}
If the input $\mb{x}^0$ is chosen so that $q^1 = q^*$, then the dynamics start at the fixed point and the distribution of $\mb{D}^l$ is independent of $l$. Moreover, even if $q^1 \ne q^*$, a few layers is often sufficient to approximately converge to the fixed point (see \cite{poole2016,schoenholz2016}). As such, when $L$ is large, it is often a good approximation to assume that $q^l = q^*$ for all depths $l$ when computing the spectrum of $\mb{J}$.

\begin{figure}[h]
  \centering
\includegraphics[width=0.8\linewidth]{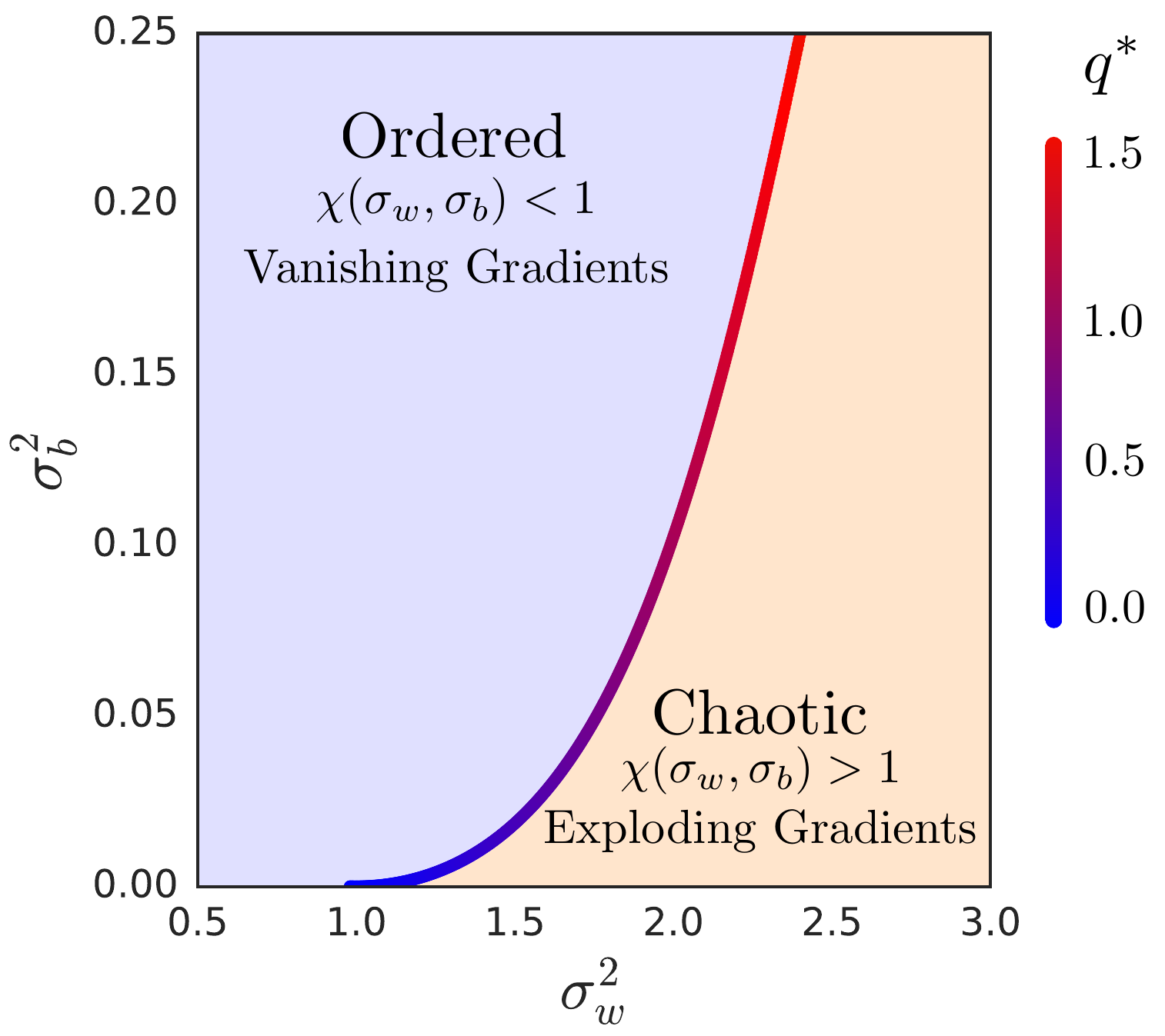}
\vspace{-.2in}
\caption{Order-chaos transition when $\phi(h) = \tanh(h)$. The critical line $\chi=1$ determines the boundary between the two phases. In the chaotic regime $\chi>1$ and gradients explode while in the ordered regime $\chi< 1$ and we expect gradients to vanish. The value of $q^*$ along this line is shown as a heatmap.}
\label{fig:tanhphasediag}
\end{figure}

Another important quantity governing signal propagation through deep networks \cite{poole2016} is 
\begin{equation}
\label{eq:chio}
\begin{split}
\chi  &= \frac{1}{N} \left\langle \text{Tr} \, (\mb{DW})^T \mb{DW} \right \rangle \\
       &= \sigma_w^2 \int \, \mathcal{D}h \left[ \phi' \left( \sqrt{q^*} h \right) \right]^2,
\end{split}
\end{equation}
where $\phi'$ is the derivative of $\phi$.  Here $\chi$ is second moment of the distribution of squared singular values of the matrix $\mb{DW}$, when the pre-activations are at their fixed point distribution with variance $q^*$.  As shown in \cite{poole2016,schoenholz2016}, $\chi(\sigma_w,\sigma_b)$ separates the $(\sigma_w,\sigma_b)$ plane into two regions: (a) when $\chi > 1$, forward signal propagation expands and folds space in a chaotic manner and back-propagated gradients exponentially explode; and (b) when $\chi < 1$, forward signal propagation contracts space in an ordered manner and back-propagated gradients exponentially vanish.   Thus the constraint $\chi(\sigma_w,\sigma_b)=1$ determines a critical line in the $(\sigma_w,\sigma_b)$ plane separating the ordered and chaotic regimes.  Moreover, the second moment of the distribution of squared singular values of $\mb{J}$ was shown simply to be $\chi^L$ in \cite{poole2016,schoenholz2016}.  Fig.~\ref{fig:tanhphasediag} shows an example of an order-chaos transition for the tanh nonlinearity.  

\subsection{Review of Free Probability}
The previous section revealed that the mean squared singular value of $\mb{J}$ is $\chi^L$. Indeed when $\chi\ll 1$ or $\chi\gg 1$ the vanishing or explosion of gradients, respectively, dominates the learning dynamics and provide a compelling case for choosing an initialization that is critical with $\chi = 1$. We would like to investigate the question of whether or not all cases where $\chi = 1$ are the same and, in particular, to obtain more detailed information about entire the singular value distribution of $\mb{J}$ when $\chi=1$. Since \eqref{eqn:Jz} consists of a product of random matrices, free probability becomes relevant as a powerful tool to compute the spectrum of $\mb{J}$, as we now review. See~\cite{mingo2017free} for a pedagogical introduction, and~\cite{pennington2017resurrecting,pennington2017geometry} for prior work applying free probability to deep learning.

In general, given a random matrix $\mb{X}$, its limiting spectral density is defined as
\begin{equation}
\rho_{X}(\lambda) \equiv \left\langle \frac{1}{N} \sum_{i=1}^N \delta(\lambda - \lambda_i) \right\rangle_X,
\end{equation}
where $\langle \cdot \rangle_X$ denotes an average w.r.t to the distribution over the random matrix $\mb{X}$.  

The \emph{Stieltjes transform} of $\rho_X$ is defined as,
\begin{equation}
\label{eqn:defg}
G_X(z) \equiv \int_{\mathbb{R}}  \frac{\rho_X(t)}{z-t}dt\,,\qquad z\in \mathbb{C}\setminus\mathbb{R}\,,
\end{equation}
which can be inverted using,
\begin{equation}
\label{eqn:inversion}
\rho_X(\lambda) = -\frac{1}{\pi} \lim_{\epsilon\to0^+} \text{Im}\, G_X(\lambda + i \epsilon)\,.
\end{equation}
$G_X$ is related to the moment generating function $M_X$,
\begin{equation}
\label{eqn:moments}
M_X(z) \equiv z G_X(z) -1 = \sum_{k=1}^{\infty} \frac{m_k}{z^k}\,,
\end{equation}
where $m_k$ is the $k$th moment of the distribution $\rho_X$,
\begin{equation}
\label{eqn:traces}
m_k  = \int d\lambda \; \rho_X(\lambda) \lambda^k = \frac{1}{N}\langle \tr \mb{X}^k \rangle_X\,.
\end{equation}
In turn, we denote the functional inverse of $M_X$ by $M_X^{-1}$, which by definition satisfies  $M_X(M_X^{-1}(z)) = M_X^{-1}(M_X(z))=z$.  
Finally, the \emph{S-transform}~\cite{speicher1994multiplicative,voiculescu1992free} is defined as,
\begin{equation}
\label{eqn:SMrelation}
S_X(z) = \frac{1+z}{z  M_X^{-1}(z)}\,.
\end{equation}
The utility of the S-transform arises from its behavior under multiplication. Specifically, if $\mb{A}$ and $\mb{B}$ are two freely independent random matrices, then the S-transform of the product random matrix ensemble $\mb{A}\mb{B}$ is simply the product of their S-transforms,
\begin{equation}
\label{eqn:Stransform}
S_{AB}(z) = S_A(z) S_B(z)\,.
\end{equation} 

\section{MASTER EQUATION FOR SPECTRAL DENSITY}

\subsection{S-transform for Jacobians}

We can now write down an implicit expression of the spectral density of $\mb{J}\mb{J}^T$, which is also the distribution of the square of the singular values of $\mb{J}$.   In particular, in the supplementary material (SM) Sec. 1, we combine \eqref{eqn:Stransform} with the facts that the S-transform depends only on traces of moments through \eqref{eqn:moments}, and that these traces are invariant under cyclic permutations, to derive a simple expression for the S-transform of $\mb{J}\mb{J}^T$,
\begin{equation}
\label{eqn:SJJT}
S_{JJ^T} =  \prod_{l=1}^{L} S_{(D^{l})^2} S_{(W^l)^T W^l} = S^L_{D^2} S^L_{W^T W}.
\end{equation}
Here the lack of dependence on the layer index $l$ on the RHS  is valid if the input $\mb{x}^0$ is such that $q^1 = q^*$. 

Thus, given expressions for the S-transforms associated with the nonlinearity, $S_{D^2}$, and the weights, $S^L_{W^T W}$, one can compute the S-transform of the input-output Jacobian $S_{JJ^T}$ at any network depth $L$ through \eqref{eqn:SJJT}.  Then from $S_{JJ^T}$, one can invert the sequence \eqref{eqn:defg}, \eqref{eqn:moments}, and \eqref{eqn:SMrelation} to obtain $\rho_{JJ^T}(\lambda)$. 

\subsection{An Efficient Master Equation}
The previous section provides a naive method for computing the spectrum $\rho_{JJ^T}(\lambda)$, through a complex sequence of calculations.  One must start from $\rho_{W^TW}(\lambda)$ and  $\rho_{D^2}(\lambda)$, compute their respective Stieltjes transforms, moment generating functions, inverse moment generating functions, and S-transforms, take the product in \eqref{eqn:SJJT}, and then invert this sequence of steps to finally arrive at $\rho_{JJ^T}(\lambda)$.  Here we provide a much simpler ``master" equation for extracting information about $\rho_{JJ^T}(\lambda)$ and its moments directly from knowledge of the moment generating function of the nonlinearity, $M_D^2(z)$, and the S-transform of the weights, $S_{W^TW}(z)$.  As we shall see, these latter two functions are the simplest functions to work with for arbitrary nonlinearities.

To derive the master equation, we insert \eqref{eqn:SMrelation}, for $\mb{X} = \mb{D}^2$, into \eqref{eqn:SJJT}, and perform some algebraic manipulations (see SM Sec.~3 for details) to obtain implicit functional equations for $M_{JJ^T}(z)$ and $G(z)$,

\begin{eqnarray}
\label{eqn:Marby}
M_{JJ^T}(z) = M_{D^2}\Big( z^{\frac{1}{L}} F\big(M_{JJ^T}(z)\big)\Big)\,,\;\;&&\\
\label{eqn:Geqn_arb}
z G(z) - 1 =  M_{D^2}\left( z^\frac{1}{L} F\big(z G(z)-1\big)\right),&&
\end{eqnarray}
where,
\begin{equation}
    F(x) = S_{W^TW}(x)\left(\frac{1+x}{x}\right)^{1-\frac{1}{L}}\,.
\end{equation}
In principle, a solution to eq.~\eqref{eqn:Geqn_arb} allows us to compute the entire spectrum of $\bm J\bm J^T$. In practice, when an exact solution in terms of elementary functions is lacking, it is still possible to extract robust numerical solutions, as we describe in the next subsection.

\begin{table*}[t!]
\caption{Properties of Nonlinearities} \label{tab:nltrans}
\begin{center}
\begin{tabular}{@{}l|l|l|l|l|l@{}}
&$\phi(h)$  & $M_{D^2}(z)$ &  $\mu_k$ & $\sigma_w^2$ &  $\sigma_{JJ^T}^2$ \\
\hline
\small{Linear}         &$h$& $\frac{1}{z-1}$  &$1$&$1$  &$L\, (-s_1)$\\ 
\small{ReLu}           &$[h]_+$&$\frac{1}{2}\frac{1}{z-1}$  &$\frac{1}{2}$&$2$  &$L\, (1-s_1)$\\
\small{Hard Tanh}     &$[h+1]_+ - [h-1]_+ - 1$& $\erf(\frac{1}{\sqrt{2 q^*}})  \frac{1}{z - 1}$  & $\erf(\frac{1}{\sqrt{2 q^*}})$& $\frac{1}{\erf(\frac{1}{\sqrt{2q^*}})}$ & $L\, (\frac{1}{\erf(\frac{1}{\sqrt{2q^*}})}-1-s_1)$ \\
\small{Erf}                 &$\erf(\frac{\sqrt{\pi}}{2}h)$& \!\!$\frac{1}{\sqrt{\pi q^*}  z}\Phi\left(\frac{1}{z}, \frac{1}{2},\frac{1+\pi q_*}{\pi q_*}\right)$ & $\frac{1}{\sqrt{1 +\pi k q_*}}$&$\sqrt{1+\pi q^*}$  & $L\, (\frac{1+\pi q^*}{\sqrt{1+2\pi q^*}}-1-s_1)$
\end{tabular}
\end{center}
\end{table*}

\subsection{Numerical Extraction of Spectra}

Here we describe how to solve \eqref{eqn:Geqn_arb} numerically. The difficulty is that \eqref{eqn:Geqn_arb} implicitly defines $G(z)$ through an equation of the form $\mathcal F(G,z)=0$. Notice that, for any given $z$, this equation may have multiple roots in $G$.  The correct branch can be chosen by requiring that $z \to \infty$,  $G(z) \sim 1/z$ \cite{TaoBook}. Therefore, one point on the correct branch can be found by taking $|z|$ large, and finding the solution to $\mathcal F(G,z)=0$ that is closest to $G = 1/z$.  Recall that to obtain the density $\rho_{JJ^T}(\lambda)$ through the inversion formula (\eqref{eqn:inversion}), we need to extract the behavior of $G(z)$ near the real axis at a point $z = \lambda + i\epsilon$ where $\rho_{JJ^T}(\lambda)$ has support. So, practically speaking, for each $\lambda$ we can walk along the imaginary direction obeying $\text{Re}(z)=\lambda$ from large imaginary values to small, and repeatedly solve $\mathcal F(G,z)=0$, always choosing the root that is closest to the previous root. 

A potential pitfall arises if we approach a point $z$ where $\mathcal F(G,z)=0$ has a double root in $G$, which could cause us to leave the correct branch of roots and then traverse an incorrect branch.  However, points in the complex two dimensional plane $(G,z) \in \mathbb{C}^2$ where $\mathcal F$ has a double root in $G$ are expected to be a set of measure $0$, and in practice they do not seem to be a concern. Algorithm~\ref{alg:root} summarizes our heuristic for computing $\rho(\lambda)$ for each $\lambda$ of interest.

\begin{algorithm}
\caption{Root finding procedure}
\label{alg:root}
\begin{algorithmic}
\STATE 1. Choose to take $2N$ steps of size $b>1$\\
\STATE 2. Initialize $z_0 = \lambda + i b^N$ and $G_0 = 1/z_0$\\
\STATE 3. For $k$ in $1\ldots 2N$:
\STATE\qquad $z_k \leftarrow \lambda + i b^{N-k}$\\
\STATE\qquad $G_k \leftarrow$ Root of \eqref{eqn:Geqn_arb} nearest to $G_{k-1}(z_k)$\\
\STATE 4. Return $-\frac{1}{\pi} \text{Im}\, G_{2N} \approx \rho(\lambda)$
\end{algorithmic}
\end{algorithm}

In the following sections, we demonstrate through many examples a precise numerical match between the outcome of Algorithm~\ref{alg:root} and direct simulations of various random neural networks, thereby justifying not only \eqref{eqn:Geqn_arb}, but also the efficacy our algorithm. 
 
\subsection{Moments of Deep Spectra}

In addition to numerically extracting the spectrum of $\mb{J}\mb{J}^T$, we can also calculate its moments $m_k$ encoded in the function
\begin{equation}
M_{JJ^T}(z) \equiv \sum_{k=1}^{\infty} \frac{m_k}{z^k}\,.
\end{equation}
These moments in turn can be computed in terms of the series expansions of $S_{W^TW}$ and $M_{D^2}$,  which we define as
\begin{eqnarray}
S_{W^TW}(z) \equiv& \sigma_w^{-2} \left(1 + \sum_{k=1}^{\infty} s_k z^k\right) \label{eqn:MomexpSt} \\
M_{D^2}(z) \equiv&  \sum_{k=1}^{\infty} \frac{\mu_k}{z^k}\,, \label{eqn:Momexps}
\end{eqnarray}
where the moments $\mu_k$  of $\mb{D}^2$ are given by,
\begin{equation}
\mu_k = \int \mathcal{D}h\; \phi'(\sqrt{q^*} h)^{2k}\,.
\end{equation}
Substituting these expansions into \eqref{eqn:Marby}, we obtain equations for the unknown moments $m_k$ in terms of the known moments $\mu_k$ and $s_k$. We can solve for the low-order moments by expanding \eqref{eqn:Marby} in powers of $z^{-1}$. By equating the coefficients of $z^{-1}$ and $z^{-2}$, we find equations for $m_1$ and $m_2$ whose solution yields (see SM Sec.~3),
\begin{equation}
\begin{split}
\label{eqn:m1m2}
m_1 &= (\sigma_w^2 \mu_1)^L\\
m_2 &= (\sigma_w^2 \mu_1)^{2L} \, L \left(\frac{\mu_2}{\mu_1^2} + \frac{1}{L} - 1 - s_1 \right)\,.
\end{split}
\end{equation}
Note the combination $\sigma_w^2 \mu_1$ is none other than $\chi$ defined in \eqref{eq:chio}, and so \eqref{eqn:m1m2} recovers the result that the mean squared singular value $m_1$ of $\mb{J}$ either exponentially explodes or vanishes unless $\chi(\sigma_w, \sigma_b)=1$ on a critical boundary between order and chaos.   However, {\it even} on this critical boundary where the mean $m_1$ of the spectrum of $\mb{J}\mb{J}^T$ is one for any depth $L$, the variance
\begin{equation}
\label{eqn:sigmaJJ}
\sigma_{JJ^T}^2 = m_2 - m_1^2 = L \left(\frac{\mu_2}{\mu_1^2} - 1 - s_1\right)
\end{equation}
grows linearly with depth $L$ for generic values of $\mu_1$, $\mu_2$ and $s_1$.
Thus $\mb{J}$ can be highly ill-conditioned at large depths $L$ for generic choices of nonlinearities and weights, even when $\sigma_w$ and $\sigma_b$  are tuned to criticality.  

\section{SPECIAL CASES OF DEEP SPECTRA}

\begin{figure*}[t]
\centering
\includegraphics[width=\linewidth]{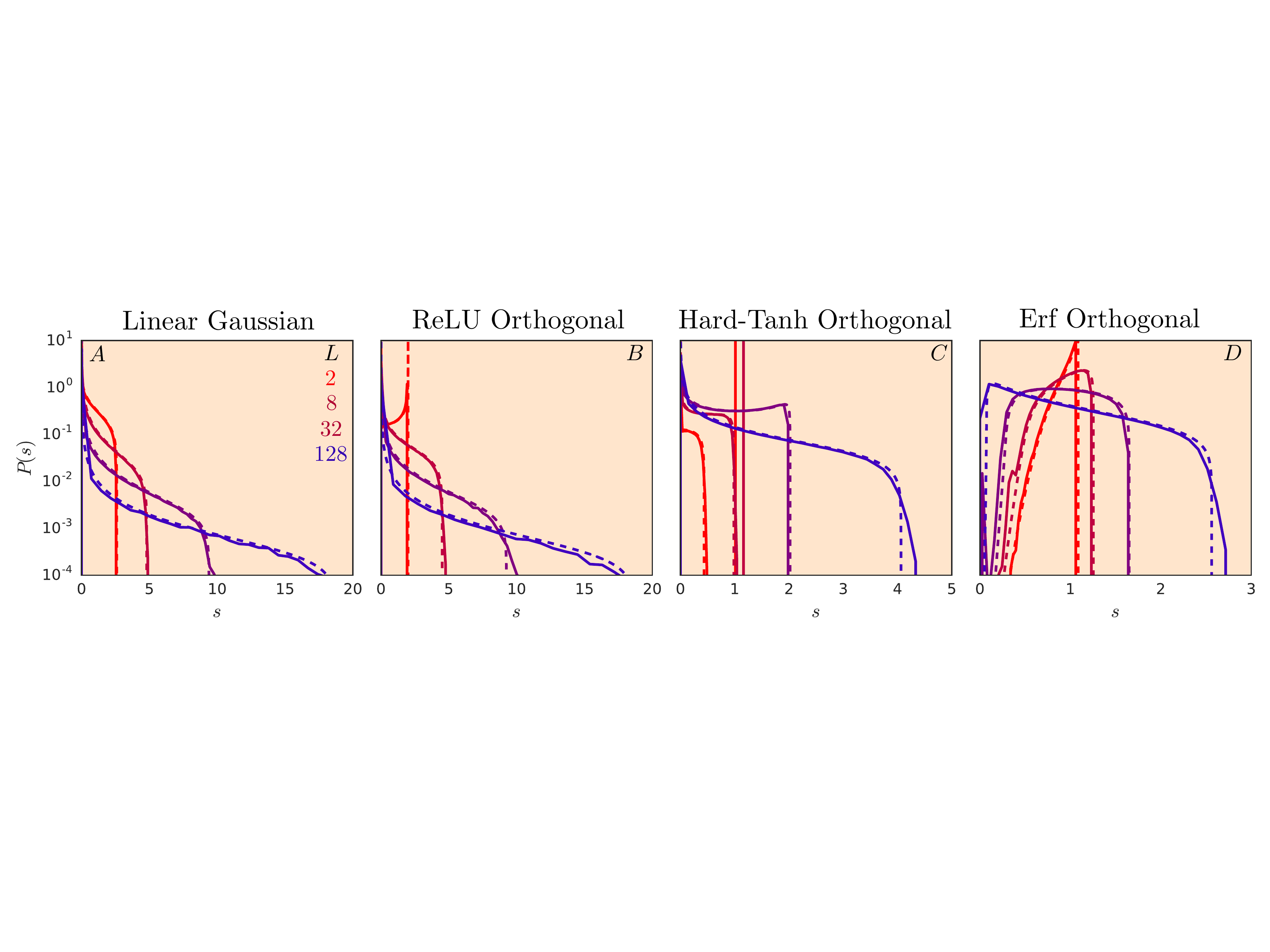}
\caption{Examples of deep spectra at criticality for different nonlinearities at different depths. Singular values from empirical simulations of networks of width 1000 are shown with solid lines while theoretical predictions from the master equation and algorithm are overlaid with dashed lines. For each panel, the weight variance $\sigma_w^2$ is held constant as the depth increases. Notice that linear Gaussian and orthogonal ReLU have similarly-shaped distributions, especially for large depths, where poor conditioning and many large singular values are observed. Erf and Hard Tanh are better conditioned, but at 128 layers we begin to observe some spread in the distributions.}
\label{fig:examplespectra}
\end{figure*}

Exploiting the master equation \eqref{eqn:Marby} requires information about $M_{D^2}(z)$, and  $S_{WW^T}(z)$.  We first provide this information and then use it to look at special cases of deep networks.

\subsection{Transforms of Nonlinearities}

First, for any nonlinearity $\phi(h)$, we have, through \eqref{eqn:defg} and \eqref{eqn:moments}, 
\begin{equation}
\label{eqn:MD2}
M_{D^2}(z) = \int \mathcal{D}h \frac{\phi'(\sqrt{q*} h)^2}{z - \phi'(\sqrt{q^*}h)^2}\,.
\end{equation}
The integral over the Gaussian measure $ \mathcal{D}h$ reflects a sum over all the activations $h^l_i$ in a layer $l$, since in the large $N$ limit the empirical distribution of activations converges to a Gaussian with standard deviation $\sqrt{q^*}$.   Moreover, an activation $h^l_i$ feels a squared slope $\phi'(h^l_i)^2$, which appears as an eigenvalue of the diagonal matrix $(\mb{D}^l)^2$.  Thus $M_{D^2}(z)$ naturally involves an integral over a function of $\phi'(\cdot)^2$ against a Gaussian.

Table \ref{tab:nltrans} provides the moment generating function and moments of $\mb{D}^2$ for several nonlinearities. Detailed derivations of the results in Table \ref{tab:nltrans}, which follow from performing the integral in \eqref{eqn:MD2}, can be found in the SM Sec.~3. In the Erf case, $\Phi$ is a special function known as the Lerch transcendent, which can be defined by its moments $\mu_k$. 

\subsection{Transforms of Weights} 
\begin{table}[h]
\caption{Transforms of weights} \label{tab:wtrans}
\begin{center}
\begin{tabular}{l|l|r@{}}
Random Matrix $\mb{W}$  & $S_{W^TW}(z)$ &$s_1$ \\
\hline
Scaled Orthogonal  & $\sigma_w^{-2}$   &$0$ \\
Scaled Gaussian     & $\sigma_w^{-2} (1+z)^{-1} $ &$-1$ \\
\end{tabular}
\end{center}
\end{table}

The S-transforms of the weights can be obtain through the sequence of equations \eqref{eqn:defg}, \eqref{eqn:moments}, and \eqref{eqn:SMrelation}, starting with $\rho_{W^TW}(\lambda) = \delta(\lambda-1)$ for an orthogonal random matrix $\mb{W}$, and $\rho_{W^TW}(\lambda) = (2\pi)^{-1}\sqrt{4-\lambda} \quad  \text{for} \, \lambda \in [0,4]$, for a Gaussian random matrix $\mb{W}$ with variance $\frac{1}{N}$ (see SM Sec.~5). Furthermore, by scaling $\mb{W} \rightarrow \sigma_w \mb{W}$, the S-transform scales as $S_{W^TW} \rightarrow \sigma_w^{-2} S_{W^TW}$, yielding the S-transforms and first moments in Table~\ref{tab:wtrans}.  

\subsection{Exact Properties of Deep Spectra}

Now for different randomly initialized deep networks, we insert the appropriate expressions in Tables \ref{tab:nltrans} and \ref{tab:wtrans} into our master equations \eqref{eqn:Marby} and \eqref{eqn:Geqn_arb} to obtain information about the spectrum of $\mb{J}\mb{J}^T$, including its entire shape, through Algorithm ~\ref{alg:root}, and its variance $\sigma_{JJ^T}^2$ through \eqref{eqn:m1m2} and \eqref{eqn:sigmaJJ}.  We always work at criticality, so that in \eqref{eq:chio}, $\chi=\sigma_w^2 \mu_1 = 1$.  The resulting condition for $\sigma_w^2$ at criticality and the value of $\sigma_{JJ^T}^2$ are shown in Table \ref{tab:nltrans} for different nonlinearities, both for orthogonal ($s_1=0$) and Gaussian ($s_1=-1$) weights.

\begin{figure*}[t]
\centering
\includegraphics[width=\linewidth]{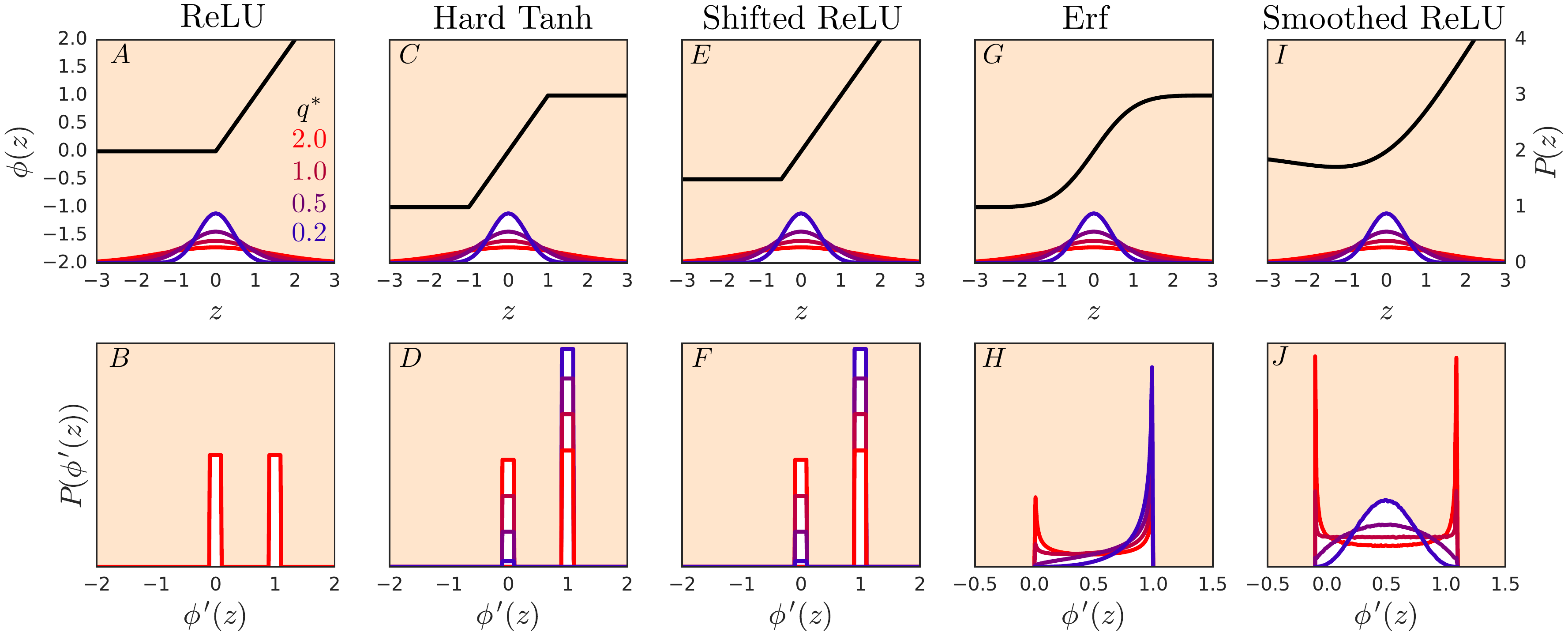}
\caption{Distribution of $\phi'(h)$ for different nonlinearities.  The top row shows the nonlinearity, $\phi(h)$, along with the Gaussian distribution of pre-activations $h$ for four different choices of the variance, $q^*$.  The bottom row gives the induced distribution of $\phi'(h)$. We see that for ReLU the distribution is independent of $q^*$. This implies that there is no stable limiting distribution for the spectrum of $\mb{JJ}^T$. By contrast for the other nonlinearities the distribution is a relatively strong function of $q^*$.}
\label{fig:gapnogap}
\end{figure*}

\subsubsection{Linear Networks}
For linear networks, the fixed point equation \eqref{eq:qliter} reduces to $q^* = \sigma_w^2 q^* + \sigma_b^2$, and $(\sigma_w, \sigma_b)=(1,0)$ is the only critical point.  Moreover, linear Gaussian networks behave very differently from orthogonal ones.  The latter are well conditioned, with $\sigma^2_{JJ^T}=0$ because the product of orthogonal matrices is orthogonal and so $\rho_{JJ^T}(\lambda) = \delta(\lambda-1)$ for all $L$.  However, $\sigma^2_{JJ^T}=L$ for Gaussian weights. This radically different behavior of the spectrum of $\mb{JJ}^T$ is shown in Fig.~\ref{fig:examplespectra}A.   

\subsubsection{ReLU Networks}
For ReLU networks, the fixed point equation \eqref{eq:qliter} reduces to $q^* = \frac{1}{2} \sigma_w^2 q^* + \sigma_b^2$, and $(\sigma_w, \sigma_b)=(\sqrt{2},0)$ is the only critical point. Unlike the linear case, $\sigma_{JJ^T}^2$ becomes $L$ for orthogonal and $2L$ for Gaussian weights.  In essence, the ReLU nonlinearity destroys the qualitative scaling advantage that linear networks possess for orthogonal weights versus Gaussian.  The qualitative similarity of spectra for ReLU Orthogonal and linear Gaussian is shown in Fig.~\ref{fig:examplespectra}AB.

\subsubsection{Hard Tanh and Erf Networks}
For Hard Tanh and Erf Networks, the criticality condition $\sigma_w^2  = {\mu_1^{-1}}$ does not determine a unique value of $\sigma_w^2$ because $\mu_1$, the mean squared slope $\phi'(h)^2$, now depends on the variance $q^*$ of the distribution of pre-activations $h$.  Since $q^*$ itself is a function of $\sigma_w$ and $\sigma_b$ through \eqref{eq:qliter}, these networks enjoy an entire critical curve in the $(\sigma_w, \sigma_b)$ plane, similar to that shown in Fig.~\ref{fig:tanhphasediag}. As $q^*$ decreases monotonically towards zero, the corresponding point on this curve approaches the point $(\sigma_w,\sigma_b)=(1,0)$.

Moreover, Table~\ref{tab:nltrans} shows that $\sigma_{JJ^T}^2 = L(\mathcal{F}(q^*) - 1 - s_1)$ with $\lim_{q^* \rightarrow 0} \mathcal{F}(q^*) = 1$.  This implies that for Gaussian weights ($s_1=-1$), no matter how small one makes $\sigma_w$, $\sigma^2_{JJ^T} \propto L$.  However, for orthogonal weights ($s_1=0$), for any fixed $L$, one can reduce $\sigma_w$ and therefore $q^*$, so as to make  $\sigma_{JJ^T}^2$ arbitrarily small. Thus Hard Tanh and Erf nonlinearities rescue the scaling advantage that orthogonal weights possess over Gaussian, which was present in linear networks, but destroyed in ReLU networks.  Examples of the well-conditioned nature of orthogonal Hard Tanh and Erf networks compared to orthogonal ReLu networks are shown in Fig.~\ref{fig:examplespectra}.

\section{UNIVERSALITY IN DEEP SPECTRA}
\begin{figure*}[t]
\centering
\includegraphics[width=\linewidth]{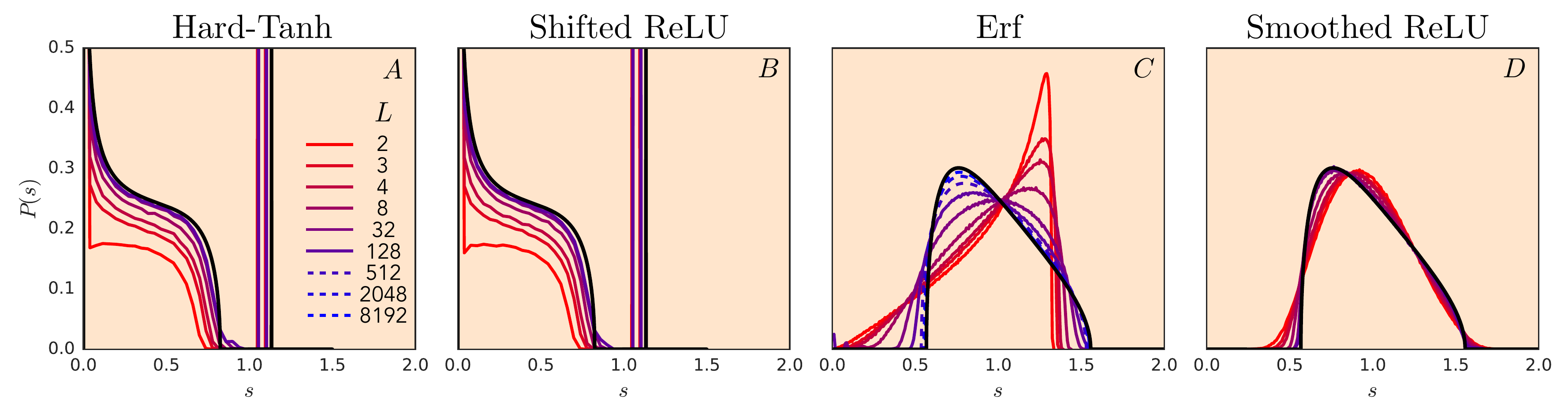}
\caption{Two limiting universality classes of Jacobian spectra. Hard Tanh and Shifted ReLU fall into one class, characterized by Bernoulli-distributed $\phi'(h)^2$, while Erf and Smoothed ReLU fall into a second class, characterized by a smooth distribution for $\phi'(h)^2$. The black curves are theoretical predictions for the limiting distributions with variance $\sigma_0^2 = 1/4$. The colored lines are emprical spectra of finite-depth width-1000 orthogonal neural networks. The empirical spectra converge to the limiting distributions in all cases. The rate of convergence is similar for Hard-Tanh and Shifted ReLU, whereas it is significantly different for Erf and Smoothed Relu, which converge to the same limiting distribution along distinct trajectories. In all cases, the solid colored lines go from shallow $L=2$ networks (red) to deep networks (purple). In all cases but Erf the deepest networks have $L=128$. For Erf, the dashed lines show solutions to \eqref{eqn:Geqn_arb} for very large depth up to $L = 8192$.}
\label{fig:univlimit}
\end{figure*}

Table~\ref{tab:nltrans} shows that for orthogonal Erf and Hard Tanh networks (but not ReLU networks), since $\sigma_{JJ^T}^2 = L(\mathcal{F}(q^*)-1)$ with $\lim_{q^* \rightarrow 0} \mathcal{F}(q^*) = 1$, one can always choose $q^*$ to vary inversely with $L$ so as to achieve a desired $L$-independent constant variance $\sigma^2_{JJ^T} \equiv \sigma^2_0$. To achieve this scaling, $q^*(L)$ should satisfy the equation $\mathcal{F}(q^*(L)) = 1 + \frac{\sigma^2_0}{L}$, which implies $\sigma_w\to 1$ and $q^* \rightarrow 0$ as $L\to\infty$.

Remarkably, in this double scaling limit, not only does the variance of the spectrum of $\mb{JJ}^T$ remain constant at the fixed value $\sigma_0^2$, but the entire {\it shape} of the distribution converges to a \emph{universal limiting distribution} as $L \rightarrow \infty$. There is more than one possible limiting distribution, but its form depends on $\phi$ only through the distribution of $\phi'(h)^2$ as $q^*\to 0$ via the expression for $M_{D^2}(z)$ in \eqref{eqn:MD2}. Therefore, many qualitatively different activation functions may in fact be members of the same \emph{universality class}. We identify two universality classes that correspond to many common activation functions: the \emph{Bernoulli} universality class and the \emph{smooth} universality class, named based on the distribution of $\phi'(h)^2$ as $q^*\to 0$.

The Bernoulli universality class contains many piecewise linear activation functions, such as Hard Tanh (Fig.~\ref{fig:gapnogap}C) and a version of ReLU shifted so as to be linear at the origin, which for concreteness we define as $\phi(x) = [x+\frac{1}{2}]_+ - \frac{1}{2}$ (Fig.~\ref{fig:gapnogap}E). While these functions look quite different, their derivatives are both Bernoulli-distributed (Fig.~\ref{fig:gapnogap}DF) and the limiting spectra of their corresponding Jacobians are the same (Fig.~\ref{fig:univlimit}AB).

The smooth universality class contains many smooth activation functions, such as Erf (Fig.~\ref{fig:gapnogap}G) and a smoothed version of ReLU that we take to be the sigmoid-weighted linear unit (SiLU)~\cite{elfwing2018sigmoid, 2017arXiv171005941R} (Fig.~\ref{fig:gapnogap}I). In this case, not only do the activation functions themselves look different, but so too do their derivatives (Fig.~\ref{fig:gapnogap}HJ). Nevertheless, in the double scaling limit, the limiting spectra of their corresponding Jacobians are the same (Fig.~\ref{fig:univlimit}CD). The rate of convergence to the limiting distribution is different, because the moments $\mu_k$ differ substantially for non-zero $q^*$.

Unlike the smoothed and shifted versions of ReLU, the vanilla ReLU activation (Fig.~\ref{fig:gapnogap}AB) behaves entirely differently and has no limiting distribution because the $\mu_k$ are independent of $q^*$ and therefore it is impossible to attain an $L$-independent constant variance $\sigma^2_{JJ^T} \equiv \sigma^2_0$ in this case.

To understand the mechanism behind the emergence of spectral universality, we now examine orthogonal networks whose activation functions have squared derivatives obeying a Bernoulli distribution and show that they all share a \emph{universal} limiting distribution as $L\to\infty$. To this end, we suppose that,
\begin{equation}
\label{eqn:MD2general}
    M_{D^2} = p(q^*)\frac{1}{z-1}\,,
\end{equation}
for some function $p(q^*)$ that measures the probability of the nonlinearity having slope one as a function of $q^*$. We will assume that $p(q^*)\to 1$ as $q^*\to 0$. The relevant ratio of moments and the weight variance $\sigma_w^2$ are given as,
\begin{equation}
\label{eqn:mugeneral}
    \frac{\mu_2}{\mu_1^2} = \frac{1}{\mu_1} = \sigma_w^2 = \frac{1}{p(q^*)}\,.
\end{equation}
From \eqref{eqn:sigmaJJ}, we have,
\begin{equation}
\label{eqn:qsJJ}
\sigma_{JJ^T}^2 = \sigma_{0}^2 = L \left(\frac{1}{p(q^*)}-1\right) \Rightarrow p(q^*) = 1 + \frac{\sigma_{0}^2}{L}\,.
\end{equation}
Notice that a solution $q^*(L)$ to \eqref{eqn:sigmaJJ} will exist for large $L$ since we are assuming $p(q^*)\to 1$ as $q^*\to 0$. Substituting this solution in \eqref{eqn:MD2general} and \eqref{eqn:mugeneral} gives for large $L$,
\begin{equation}
    M_{D^2} = \frac{L}{L+\sigma_0^2}\frac{1}{z-1}\quad\text{and}\quad \mu_1 = \frac{L}{L+\sigma_0^2}\,.
\end{equation}
Using these expressions and \eqref{eqn:SMrelation}, we find that the S-transform obeys,
\begin{equation}
S^{\text{Bernoulli}}_{JJ^T} = (\mu_1 \frac{1+z}{z M^{-1}_{D^2}})^L = \left(1 + \frac{z \sigma_{0}^2}{L(1+z)}\right)^{-L}\,.
\end{equation}
The large depth limit gives,
\begin{equation}
S^{\text{Bernoulli}}_{JJ^T}  = e^{-\frac{z \sigma_{0}^2}{(1+z)}}\,.
\end{equation}
Using \eqref{eqn:moments} and \eqref{eqn:SMrelation} to solve for $G(z)$ gives,
\begin{equation}
\label{eqn:BernoulliG}
G(z) = \frac{1}{z} \frac{\sigma_{0}^2}{\sigma_{0}^2+W(\frac{-\sigma_{0}^2}{z})}\,,
\end{equation}
where $W$ denotes the principal branch of the Lambert-W function~\cite{corless1996lambertw} and solves the transcendental equation,
\begin{equation}
    W(x) e^{W(x)} = x\,.
\end{equation}
The spectral density can be extracted from \eqref{eqn:BernoulliG} easily using \eqref{eqn:inversion}. The results are shown in black lines in Fig.~\ref{fig:univlimit}AB. Both Hard Tanh and Shifted ReLU have Bernoulli-distributed $\phi'(h)^2$ and, despite being qualitatively different activation functions, have the same limiting spectral distributions. It is evident that the empirical spectral densities converge to this universal limiting distribution as the depth increases.

Next we build some additional understanding of the spectral density implied by \eqref{eqn:BernoulliG}. Because the spectral density is proportional to the imaginary part of $G(z)$, we expect the locations of the spectral edges to be related to branch points of $G(z)$, or more generally to poles in its derivative. Using the relation,
\begin{equation}
W'(x) = \frac{1}{x + e^{W(x)}}\,,
\end{equation}
we can inspect the derivative of $G(z)$. It may be expressed as,
\begin{equation}
\label{eqn:BernoulliGprime}
G'(z) = -\frac{\sigma_0^2\big(\sigma_0^2 +W(\frac{-\sigma_{0}^2}{z})(\sigma_0^2 +W(\frac{-\sigma_{0}^2}{z})\big)}{z^2\big(1+W(\frac{-\sigma_{0}^2}{z})\big)\big(\sigma_0^2 + W(\frac{-\sigma_{0}^2}{z})\big)^2}\,.
\end{equation}
By inspection, we find that $G'(z)$ has double poles at,
\begin{equation}
z = \lambda_{0} = 0, \quad z = \lambda_{2} = e^{\sigma_{0}^2}\,,
\end{equation}
which are locations where the spectral density diverges, i.e. there are delta function peaks at $\lambda_0$ and $\lambda_2$. Note that there is only a pole at $\lambda_2$ if $\sigma_0 \le 1$. There is also a single pole at,
\begin{equation}
\lambda_1 = \sigma_{0}^2 e\,,
\end{equation}
which defines the right spectral edge, i.e. the maximum value of the bulk of the density.

The above observations regarding $\lambda_0$, $\lambda_1$, and $\lambda_2$ are evident in Fig.~\ref{fig:univlimit}AB. Noting that in the figure, $\sigma_0 = 1/2$, we predict that the bulk of the density to have its right edge located at $s = \sqrt{\lambda_1} = \sqrt{e}/2 \approx 0.82$ and that there should be a delta function peak at $s=\sqrt{\lambda_2} = e^{1/8} \approx 1.13$, both of which are reflected in the figure.

A similar analysis can be carried out for activation functions for which the distribution of $\phi'(h)^2$ is smooth and concentrates around one as $q^*\to 0$. The analysis for Erf is presented in the SM. We find that,
\begin{equation}
S^{\text{Smooth}}_{JJ^T}  = e^{-z \sigma_0^2}\,,
\end{equation}
and that $G(z)$ can be expressed in terms of a generalized Lambert-W function~\cite{mezHo2017generalization}. The locations of the spectral edges are given by $s_{\pm} = e^{-\frac{1}{4}\sigma_\pm^2}\sqrt{1 + \frac{1}{2}\sigma_\mp^2}$, where,
\begin{equation}
\sigma_\pm^2 = \sigma_0\Big(\sigma_0\pm \sqrt{\sigma_0^2 + 4}\Big)\,.
\end{equation}
For $\sigma_0=1/2$, these results give $s_{-}\approx0.57$ and $s_+=1.56$, which is in excellent agreement with the behavior observed in Fig.~\ref{fig:univlimit}CD. Overall, Fig.~\ref{fig:univlimit} provides strong evidence supporting our predictions that orthogonal Hard Tanh and shifted ReLU networks have the Bernoulli limit distribution, while orthogonal Erf and smoothed Relu networks have the smooth limit distribution.  

Finally, we derived these universal limits assuming orthogonal weights. In the SM we show that orthogonality is in fact necessary for the existence of a stable limiting distribution for the spectrum of $\mb{JJ}^T$.  No other random matrix ensemble can yield a stable distribution for {\it any} choice of nonlinearity with $\phi'(0)=1$. Essentially, any spread in the singular values of $\mb{W}$ grows in an unbounded way with depth and cannot be nonlinearly damped.

\section{DISCUSSION}
In summary, motivated by a lack of theoretical clarity on when and why different weight initializations and nonlinearities combine to yield well-conditioned spectra that speed up deep learning, we developed a calculational framework based on free probability to provide, with unprecedented detail, analytic information about the {\it entire} Jacobian spectrum of deep networks with {\it arbitrary} nonlinearities. Our results provide a principled framework for the initialization of weights and the choice of nonlinearities in order to produce well-conditioned Jacobians and fast learning. Intriguingly, we find novel universality classes of deep spectra that remain well-conditioned as the depth goes to infinity, as well as theoretical conditions for their existence. Our results lend additional support to the surprising conclusions revealed in~\cite{pennington2017resurrecting}, namely that using either Gaussian initializations or ReLU nonlinearities precludes the possibility of obtaining stable spectral distributions for very deep networks. Beyond the sigmoidal units advocated in~\cite{pennington2017resurrecting}, our results suggest that a wide variety of nonlinearities, including shifted and smoothed variants of ReLU, can achieve dynamical isometry, provided the weights are orthogonal.  Interesting future work could involve the discovery of new universality classes of well-conditioned deep spectra for more diverse nonlinearities than considered here.
\bibliography{DeepSpectra}
\bibliographystyle{unsrt}

\onecolumn

\newpage
\begin{center}
\textbf{\Large The Emergence of Spectral Universality in Deep Networks:\\Supplementary Material}
\end{center}
\setcounter{equation}{0}
\setcounter{figure}{0}
\setcounter{table}{0}
\setcounter{page}{1}
\setcounter{section}{0}
\makeatletter
\renewcommand{\theequation}{S\arabic{equation}}
\renewcommand{\thefigure}{S\arabic{figure}}

\section{Review of free probability}

For what follows, we define the key objects of free probability.  Given a random matrix $\mb{X}$, its limiting spectral density is defined as
\begin{equation}
\rho_{X}(\lambda) \equiv \left\langle \frac{1}{N} \sum_{i=1}^N \delta(\lambda - \lambda_i) \right\rangle_X,
\end{equation}
where $\langle \cdot \rangle_X$ denotes an average w.r.t to the distribution over the random matrix $\mb{X}$.  For large $N$, the empirical histogram of eigenvalues of a single realization of $\mb{X}$ converges to $\rho_X$.  In turn, the \emph{Stieltjes transform} of $\rho_X$ is defined as,
\begin{equation}
G_X(z) \equiv \int_{\mathbb{R}}  \frac{\rho_X(t)}{z-t}dt\,,\qquad z\in \mathbb{C}\setminus\mathbb{R}\,,
\end{equation}
which can be inverted using,
\begin{equation}
\label{eqn:inversionS}
\rho_X(\lambda) = -\frac{1}{\pi} \lim_{\epsilon\to0^+} \text{Im}\, G_X(\lambda + i \epsilon)\,.
\end{equation}
$G_X$ is related to the moment generating function $M_X$,
\begin{equation}
\label{eqn:momentsS}
M_X(z) \equiv z G_X(z) -1 = \sum_{k=1}^{\infty} \frac{m_k}{z^k}\,,
\end{equation}
where the $m_k$ is the $k$'th moment of the distribution $\rho_X$,
\begin{equation}
\label{eqn:tracesS}
m_k  = \int d\lambda \; \rho_X(\lambda) \lambda^k = \frac{1}{N}\langle \tr \mb{X}^k \rangle_X\,.
\end{equation}
In turn, we denote the functional inverse of $M_X$ by $M_X^{-1}$, which by definition satisfies  $M_X(M_X^{-1}(z)) = M_X^{-1}(M_X(z))=z$.  
Finally, the \emph{S-transform}~\cite{speicher1994multiplicative,voiculescu1992free} is defined in terms of the functional inverse $M_X^{-1}$ as,
\begin{equation}
\label{eqn:SMrelationS}
S_X(z) = \frac{1+z}{z  M_X^{-1}(z)}\,.
\end{equation}
The utility of the S-transform arises from its behavior under multiplication. Specifically, if $\mb{A}$ and $\mb{B}$ are two freely independent random matrices, then the S-transform of the product random matrix ensemble $\mb{A}\mb{B}$ is simply the product of their S-transforms,
\begin{equation}
\label{eqn:StransformS}
S_{AB}(z) = S_A(z) S_B(z)\,.
\end{equation}

\section{Free probability and deep networks}

We will now use eqn.~(\ref{eqn:StransformS}) to write down an implicit definition of the spectral density of $\mb{J}\mb{J}^T$, which is also the distribution of the square of the singular values of $\mb{J}$.  Here $\mb{J}$ is the input-output Jacobian of a deep network defined in the main paper.  First notice that, by eqn.~(\ref{eqn:moments}), $M(z)$ and thus $S(z)$ depend only on the moments of the spectral density. The moments, in turn, can be defined in terms of traces (as in eqn. (\ref{eqn:tracesS})), which are invariant to cyclic permutations, i.e.,
\begin{equation}
\tr (\A^1 \A^2\cdots \A^m)^k = \tr (\A^2\cdots \A^m \A^1)^k\,.
\end{equation}
Therefore the S-transform is invariant to cyclic permutations. Now define matrices $\mb{Q}^l$ and $\tilde{\mb{Q}}^l$ as,
\begin{equation}
\begin{split}
\mb{Q}^L &\equiv \mb{J}\mb{J}^T = ( \mb{D}^{L} \mb{W}^L \mb{D}^{L-1} \cdots \mb{D}^1 \mb{W}^1)(\mb{D}^{L} \mb{W}^L \mb{D}^{L-1}\cdots \mb{D}^1 \mb{W}^1)^T\\
\tilde{\mb{Q}}^L &\equiv  \left[ (\mb{D}^{L} \mb{W}^{L})^T \mb{D}^{L} \mb{W}^L \right](\mb{W}^{L-1} \mb{D}^{L-2}\cdots \mb{D}^1 \mb{W}^1) (\mb{W}^{L-1} \mb{D}^{L-2} \cdots \mb{D}^1 \mb{W}^1)^T \\
& =  \left[ (\mb{D}^{L} \mb{W}^{L})^T \mb{D}^{L} \mb{W}^L \right] \mb{Q}^{L-1}\,.
\end{split}
\end{equation}
Now $\mb{Q}^L$ and  $\tilde{\mb{Q}}^L$ are related by a cyclic permutation. Therefore the above argument shows that their S-transforms are equal, i.e. $S_{Q_L} = S_{\tilde{Q}_L}$. Furthermore $(\mb{D}^{L} \mb{W}^{L})^T \mb{D}^{L} \mb{W}^L$  and $(\mb{D}^{L})^2 (\mb{W}^{L})^T \mb{W}^L  $ are related by a cyclic permutation, implying their S-transforms are also equal.   Then a recursive application of eqn.~(\ref{eqn:StransformS}) and cyclic invariance of S-transforms implies that,
\begin{equation}
\label{eqn:SJJTS}
S_{JJ^T} = S_{Q_L}  = S_{(D^{L})^2} S_{(W^{L})^T W^L} S_{Q^{L-1}} = \prod_{l=1}^{L} S_{(D^{l})^2} S_{(W^l)^T W^l} = S^L_{D^2} S^L_{W^T W}\\
\end{equation}
where the last equality follows if each term in the Jacobian product identically distributed. \\
\\
Given the expression for $S_{JJ^T}$, a simple procedure recovers the density of singular values of $\mb{J}$: 
\begin{enumerate}
\item Use eqn.~(\ref{eqn:SMrelationS}) to obtain the moment generating function $M_{JJ^T}(z)$
\item Use eqn.~(\ref{eqn:moments}) to obtain the Stieltjes transform $G_{JJ^T}(z)$
\item Use eqn.~(\ref{eqn:inversionS}) to obtain the spectral density $\rho_{JJ^T}(\lambda)$
\item Use the relation $\lambda = \sigma^2$ to obtain the density of singular values of $J$.
\end{enumerate}
So in order to compute the distribution of singular values of of $J$, all that remains is to compute the S-transforms of $W^TW$ and of $D^2$. We will attack this problem for specific activation functions and matrix ensembles in the following sections.

\section{Derivation of master equations for the spectrum of the Jacobian}

To derive the master equation, we first insert \eqref{eqn:SMrelationS}, for $\mb{X} = \mb{D^2}$, into \eqref{eqn:SJJTS} to obtain
\begin{equation*}
\label{eqn:temp1}
S_{JJ^T} = S_{W^TW}^L\left(\frac{1+z}{z}\right)^{L}\left(M_{D^2}^{-1}\right)^{-L}\,. 
\end{equation*}
Then we find $M_{JJ^T}^{-1} = (1+z)(zS_{J^TJ})^{-1}$ by inverting \eqref{eqn:SMrelationS}, which combined with the above equation yields
\begin{equation*}
M_{JJ^T}^{-1} = S_{W^TW}^{-L}\left(\frac{1+z}{z}\right)^{1-L}\left(M_{D^2}^{-1}\right)^{L}\,.
\end{equation*}
Then solving for $M_{D^2}^{-1}$ yields
\begin{equation*}
M_{D^2}^{-1}  = \left(M_{JJ^T}^{-1} S_{W^TW}^{L}\left(\frac{1+z}{z}\right)^{L-1}\right)^{\frac{1}{L}}\,.
\end{equation*}
Applying $M_{D^2}$ to both sides gives,
\begin{equation*}
z  = M_{D^2}\left( \Big(M_{JJ^T}^{-1} S_{W^TW}^{L}(z)\left(\frac{1+z}{z}\right)^{L-1}\Big)^{\frac{1}{L}}\right)\,.
\end{equation*}
Finally, evaluating this equation at $z = M_{JJ^T}$ gives our sought after master equation:
\begin{equation}
\label{eqn:MarbyS}
M_{JJ^T}(z) = M_{D^2}\left( z^{\frac{1}{L}}\,S_{W^TW}\big(M_{JJ^T}(z)\big)\, \left(1+\frac{1}{M_{JJ^T}(z)}\right)^{1-\frac{1}{L}}\right)\,.
\end{equation}
This is an implicit functional equation for $M_{JJ^T}(z)$, an unknown quantity, in terms of the known functions $M_{D^2}(z)$ and $S_{W^TW}(z)$.   
Furthermore, by substituting \eqref{eqn:momentsS}, $M_{JJ^T} = zG_{JJ^T}-1$, into \eqref{eqn:MarbyS}, we also obtain an implicit functional equation for the Stieltjes transform $G$ of $\rho_{JJ^T}(\lambda)$,
\begin{equation}
\label{eqn:Geqn_arbS}
z G - 1  =M_{D^2}\left( z^\frac{1}{L} S_{W^TW}(z G-1)\left(\frac{z G}{z G - 1}\right)^{1-\frac{1}{L}}\right).
\end{equation}

\section{Derivation of Moments of deep spectra}
The moments $m_k$ of  the spectrum of $\mb{J}\mb{J}^T$ are encoded in the moment generating function
\begin{equation}
M_{JJ^T}(z) \equiv \sum_{k=1}^{\infty} \frac{m_k}{z^k}\,.
\end{equation}
These moments in turn can be computed in terms of the series expansions of $S_{W^TW}$ and $M_{D^2}$,  which we define as
\begin{eqnarray}
S_{W^TW}(z) \equiv& \sigma_w^{-2} \left(1 + \sum_{k=1}^{\infty} s_k z^k\right) \label{eqn:MomexpStS} \\
M_{D^2}(z) \equiv&  \sum_{k=1}^{\infty} \frac{\mu_k}{z^k}\,, \label{eqn:MomexpsS}
\end{eqnarray}
where the moments $\mu_k$  of $\mb{D}^2$ are given by,
\begin{equation}
\mu_k = \int \mathcal{D}h\; \phi'(\sqrt{q^*} h)^{2k}\,.
\end{equation}

We can substitute these moment expansions into \eqref{eqn:MarbyS} to obtain equations for the unknown moments $m_k$ of the spectrum of $\mb{J}\mb{J}^T$, in terms of the known moments $\mu_k$ and $s_k$.  We can solve for the low order moments by expanding  \eqref{eqn:MarbyS} in powers of $z^{-1}$.  By equating the coefficients of $z^{-1}$ and $z^{-2}$, we obtain the following equations for $m_1$ and $m_2$,
\begin{equation}
\begin{split}
m_1 &= \sigma_w^2 \mu_1 m_1^{1-\frac{1}{L}}\\
m_2 &= \sigma_w^4 \mu_2 m_1^{2-\frac{2}{L}} \\
        &+ \sigma_w^2 \mu_1 m_1^{2-\frac{1}{L}}\left(\Big(\frac{m_2}{m_1^2}-1\Big)\Big(1-\frac{1}{L}\Big) - s_1\right)\,.
\end{split}
\end{equation}
Solving for $m_1$ and $m_2$ yields,
\begin{equation}
\begin{split}
\label{eqn:m1m2S}
m_1 &= (\sigma_w^2 \mu_1)^L\\
m_2 &= (\sigma_w^2 \mu_1)^{2L} \, L \left(\frac{\mu_2}{\mu_1^2} + \frac{1}{L} - 1 - s_1 \right)\,.
\end{split}
\end{equation}

\section{Transforms of Nonlinearities}
Here we compute the moment generating functions $M_{D^2}(z)$ for various choices of the nonlinearity $\phi$, some of which are displayed in Table 1 of the main paper. 
\subsection{$\phi(x) = x$}
\begin{equation}
\begin{split}
M_{D^2}(z) &= \int \mathcal{D}x \frac{1}{z - 1}\\
&= \frac{1}{z-1}\,.
\end{split}
\end{equation}
\subsection{$\phi(x) = [x]_+$}
\begin{equation}
\begin{split}
M_{D^2}(z) &= \int \mathcal{D}x \frac{\theta(x)^2}{z - \theta(x)^2}\\
&= \frac{1}{2} \int \mathcal{D}x \frac{1}{z - 1}\\
&= \frac{1}{2} \frac{1}{z - 1}\,.
\end{split}
\end{equation}
\subsection{$\phi(x) = \htanh(x)$}
\begin{equation}
\begin{split}
M_{D^2}(z) &= \int \mathcal{D}x \frac{\theta(1-q_* x)^2}{z - \theta(1+ q_* x)^2}\\
&= \erf(\frac{1}{\sqrt{2} q_*}) \int \mathcal{D}x \frac{1}{z - 1}\\
&= \erf(\frac{1}{\sqrt{2} q_*})  \frac{1}{z - 1}\,.
\end{split}
\end{equation}
\subsection{$\phi(x) = [x]_+ + \alpha [-x]_+$}
\begin{equation}
\begin{split}
M_{D^2}(z) &= \int \mathcal{D}x \frac{\phi'(q_* x)^2}{z - \phi'(q_* x)^2}\\
&=\frac{1}{2(z-1)} + \frac{1}{2(z/\alpha^2-1)}\,.
\end{split}
\end{equation}
\subsection{$\phi(x) = \erf(\frac{\sqrt{\pi}}{2}x)$}
\begin{equation}
\label{eqn:erf_M}
\begin{split}
M_{D^2}(z) &= \int \mathcal{D}x \frac{\phi'(q_* x)^2}{z - \phi'(q_* x)^2}\\
&= \sum_{k=1}^{\infty}z^{-k} \int \mathcal{D}x\; e^{-\frac{1}{2}\pi k q_*^2 x^2}\\
&= \sum_{k=1}^{\infty}\frac{1}{z^k \sqrt{1 +\pi k q_*^2}}\\
&= \frac{1}{\sqrt{\pi} q_* z}\Phi\left(\frac{1}{z}, \frac{1}{2}, 1 + \frac{1}{\pi q_*^2}\right)\,,
\end{split}
\end{equation}
where $\Phi$ is the special function known as the Lerch transcendent. 
\subsection{$\phi(x) = \frac{2}{\pi}\arctan(\frac{\pi}{2}x)$}
\begin{equation}
\begin{split}
M_{D^2}(z) &= \int \mathcal{D}x \frac{\phi'(q_* x)^2}{z - \phi'(q_* x)^2}\\
&= \int \mathcal{D}x\; \frac{16}{(4+\pi^2 q_*^2 x^2)^2 z + 16}\\
&= -\frac{\sqrt{2}}{\pi^{3/2} q_*^2 \sqrt{z}}\left(\frac{e^{\frac{z_+}{2}}}{\sqrt{z_+}}\erfc\left(\sqrt{\frac{z_+}{2}}\right) - \frac{e^{\frac{z_-}{2}}}{\sqrt{z_-}}\erfc\left(\sqrt{\frac{z_-}{2}}\right)\right)\,,
\end{split}
\end{equation}
where,
\begin{equation}
z_{\pm} = \frac{4(\sqrt{z} \pm 1)}{\pi^2 q_*^2 \sqrt{z}}\,.
\end{equation}

\section{Transforms of Weights}

First consider the case of an orthogonal random matrix satisfying $\mb{W^TW} = \mb{I}$.  Then 
\begin{equation}
\begin{split}
\rho_{W^TW}(\lambda) &= \delta(\lambda-1) \\
G_{W^TW}(z) &= (z-1)^{-1} \\ 
M_{W^TW}(z) &= (z-1)^{-1} \\
M_{W^TW}^{-1}(z) &= (1+z)/z \\
S_{W^TW}(z) &= 1. \\
\end{split}
\end{equation}
The case of a random Gaussian random matrix $\mb{W}$ with zero mean, variance $\frac{1}{N}$ entries is more complex, but well known:
\begin{equation}
\begin{split}
\rho_{W^TW}(\lambda) &= (2\pi)^{-1}\sqrt{4-\lambda} \quad  \text{for} \, \lambda \in [0,4] \\
G_{W^TW}(z) &= \frac{1}{2}\left(1- \sqrt{\frac{z-4}{2}}\right) \\ 
M_{W^TW}(z) &=  \frac{1}{2}(z-\sqrt{z(z-4)}-2) \\
M_{W^TW}^{-1}(z) &= (1+z)^2/z \\
S_{W^TW}(z) &= (1+z)^{-1}. \\
\end{split}
\end{equation}
Furthermore, by scaling $\mb{W} \rightarrow \sigma_w \mb{W}$, the S-transform scales as $S_{W^TW} \rightarrow \sigma_w^{-2} S_{W^TW}$, yielding the S-transforms in Table \ref{tab:wtransS}.  
\begin{table}[h]
\caption{Transforms of weights} \label{tab:wtransS}
\begin{center}
\begin{tabular}{l|l|r}
Random Matrix $\mb{W}$  & $S_{W^TW}(z)$ &$s_1$ \\
\hline \\
Scaled Orthogonal  & $\sigma_w^{-2}$   &$0$ \\
Scaled Gaussian     & $\sigma_w^{-2} (1+z)^{-1} $ &$-1$ \\
\end{tabular}
\end{center}
\end{table}

\section{Universality class of orthogonal Hard Tanh networks}

We consider hard tanh with orthogonal weights. The moment generating function is,
\begin{equation}
M_{D^2} = \erf(\frac{1}{\sqrt{2q^*} })  \frac{1}{z - 1}\,,
\end{equation}
so that
\begin{equation}
\frac{\mu_2}{\mu_1^2} = \frac{1}{\erf(\frac{1}{\sqrt{2q^*}})}\,\quad\text{and}\quad g = \frac{1}{\sqrt{\mu_1}} = \frac{1}{\erf(\frac{1}{\sqrt{2q^*}})}\,.
\end{equation}
Also we have,
\begin{equation}
\sigma_{JJ^T}^2 = L\,\big(\frac{1}{\erf(\frac{1}{\sqrt{2q^*}})}-1\big) \quad \Rightarrow \quad q_*(L) = \frac{1}{\sqrt{2 \erf^{-1}(\frac{L}{L+\sigma_{0}^2})}}\,.
\end{equation}
if we wish to scale $q^*$ with depth $L$ so as to achieve a depth independent constant variance $\sigma_{JJ^T}^2 = \sigma_0^2$ as $L \rightarrow \infty$. 
This expression for $q^*$ gives,
\begin{equation}
M_{D^2} = \frac{L}{L+\sigma_{0}^2}\frac{1}{z - 1}\quad\text{and}\quad \mu_1 = \frac{L}{L+\sigma_{0}^2}\,,
\end{equation}
so that,
\begin{equation}
S_{JJ^T} = S_{D^2} = (\mu_1 \frac{1+z}{z M^{-1}_{D^2}})^L = \left(\frac{L(1+z)}{L(1+z) + z \sigma_{0}^2}\right)^L = \left(1 + \frac{z \sigma_{0}^2}{L(1+z)}\right)^{-L}\,.
\end{equation}
The large depth limit gives,
\begin{equation}
S_{JJ^T}  = e^{-\frac{z \sigma_{0}^2}{(1+z)}}\,.
\end{equation}
Solving for $G(z)$ gives,
\begin{equation}
G(z) = \frac{1}{z} \frac{1}{1+W(\frac{-\sigma_{0}}{z})/\sigma_{0}^2}\,,
\end{equation}
where $W$ is the standard Lambert-W function, or product log. The derivative of this function has double poles at,
\begin{equation}
\lambda_{0} = 0, \quad \lambda_{2} = e^{\sigma_{0}^2}\,,
\end{equation}
which are locations where the spectral density diverges. There is also a single pole at,
\begin{equation}
\lambda_1 = \sigma_{0}^2 e\,,
\end{equation}
which is the maximum value of the bulk of the density.

\section{Universality class of orthogonal $\erf$ networks}
Consider $\phi(x) = \sqrt{\frac{\pi}{2}}\erf(\frac{x}{\sqrt{2}}),$ which has been scaled so that $\phi'(0) = 1$ and $\phi'''(0) = -1$. The $\mu_k$ are given by,
\begin{equation}
\mu_k = \frac{1}{\sqrt{1 + 2 k q^*}}\,
\end{equation}
so that
\begin{equation}
\label{eqn:sigma2}
\sigma_{JJ^T}^2 = L\, \Big(\frac{1 + 2 q^*}{\sqrt{1 +4 q^*}} - 1 \Big)\,.
\end{equation}
If we wish to scale $q^*$ with depth $L$ so as to achieve a depth independent constant variance $\sigma_{JJ^T}^2 = \sigma_0^2$ as $L \rightarrow \infty$, then we can choose 

\begin{equation}
q^*(L) = \left({\frac{\sigma_0^2}{L}+\frac{\sigma_0^4}{2 L^2}+\left(\frac{\sigma_0}{2L}+\frac{\sigma_0^3}{2L^2}\right)\sqrt{2L+\sigma_0^2}}\right)^{1/4}\,.
\end{equation}
Since we also assume the network is critical, we also have that,
\begin{equation}
\sigma_w^2 = (1 + 2 q^*)^\frac{1}{4}\,.
\end{equation}
\\
To illustrate universality, we next consider an arbitrary activation function, and assume that it has a Taylor expansion around 0. This allows us to expand the $\mu_k$. First we write,
\begin{equation}
\phi(x) = \sum_{k=0}^{\infty} \phi_k x^k\,,
\end{equation}
We will need $\phi_1 \ne 0$. First we will assume that  $\phi_2 \ne 0$. Using this expansion we can write,
\begin{equation}
\mu_k = \phi_1^{2k} \left(1 + k\Big((2k-1)\frac{\phi_2^2}{\phi_1^2} + \frac{\phi_3}{\phi_1}\Big)q_*^2 + \mathcal{O}(q_*^4)\right)\,.
\end{equation}
We also have
\begin{equation}
\begin{split}
S_{JJ^T} & = \left(\mu_1 \frac{1+z}{z M_{D^2}^{-1}}\right)^{L}\,,\\
\end{split}
\end{equation}
where we have used the fact that the network is critical so that we have $\mu_1 = g^{-2}$. Using the Lagrange inversion theorem to expand $M_{D^2}^{-1}$, we find that 
\begin{equation}
\mu_1 \frac{1+z}{z M_{D^2}^{-1}} = 1 - 4\frac{\phi_2^2}{\phi_1^2}z q_*^2 + \mathcal{O}(q_*^4)\,.
\end{equation}
Meanwhile,
\begin{equation}
\begin{split}
L &= \sigma_0^2 \,\Big(\frac{\mu_2}{\mu_1^2}-1\Big)^{-1},\\
&= \sigma_0^2 \frac{\phi_1^2}{4\phi_2^2 q_*^2}\,,
\end{split}
\end{equation}
so that,
\begin{equation}
\begin{split}
S_{JJ^T} & = \left(\mu_1 \frac{1+z}{z M_{D^2}^{-1}}\right)^{L}\\
&=\Big(1 - \frac{\sigma_0^2}{L} z\Big)^L \\
&= e^{-\sigma_0^2 z} + \mathcal{O}(L^{-1})\,,
\end{split}
\end{equation}
Next we will assume that  $\phi_2 = 0$ and $\phi_3 \ne 0$\footnote{We suspect these additional assumptions are unnecessary and that the results which follow are valid so long as there exists a $k$ for which $\phi_k \ne 0$. It would be interesting to prove this.}. Using the above expansion we can write,
\begin{equation}
\mu_k = \phi_1^{2k} \left(1 + k\frac{\phi_3}{\phi_1}q_*^2 + \mathcal{O}(q_*^4)\right)\,.
\end{equation}
Also we have,
\begin{equation}
\begin{split}
S_{JJ^T} & = \left(\mu_1 \frac{1+z}{z M_{D^2}^{-1}}\right)^{L}\,,\\
\end{split}
\end{equation}
where we have used the fact that the network is critical so that we have $\mu_1 = \sigma_w^{-2}$. Using the Lagrange inversion theorem to expand $M_{D^2}^{-1}$, we find that 
\begin{equation}
\mu_1 \frac{1+z}{z M_{D^2}^{-1}} = 1 - 2\frac{\phi_3^2}{\phi_1^2}z q_*^4 + \mathcal{O}(q_*^6)\,.
\end{equation}
Meanwhile,
\begin{equation}
\begin{split}
L &= \sigma^2 \,\Big(\frac{\mu_2}{\mu_1^2}-1\Big)^{-1},\\
&= \sigma^2 \frac{\phi_1^2}{2\phi_3^2 q_*^4}\,,
\end{split}
\end{equation}
so that,
\begin{equation}
\begin{split}
S_{JJ^T} & = \left(\mu_1 \frac{1+z}{z M_{D^2}^{-1}}\right)^{L}\\
&=\Big(1 - \frac{\sigma_0^2}{L} z\Big)^L \\
&= e^{-\sigma_0^2 z} + \mathcal{O}(L^{-1})\,,
\end{split}
\end{equation}
establishing a universal limiting S-transform (subject to our assumptions). From this result we can extract the Stieltjes transform and thus the spectral density. The result establishes a universal double scaling limiting spectral distribution.\\
\\
Next we observe that the Stieltjes transform can be expressed in terms of a generalization of the Lambert -$W$ function called the r-Lambert function, $W_r(z)$, which is defined by
\begin{equation}
\label{eqn:Wr}
W_r e^{W_r} + r W_r = z\,.
\end{equation}
In terms of this function, the Stieltjes transform is,
\begin{equation}
G(z) = \frac{W_{-e^{\sigma_0^2}z}(-\sigma_0^2 z e^{\sigma_0^2})}{z \sigma_0^2}\,.
\end{equation}
We can extract the maximum and minumum eigenvalue by finding the branch points of this function. It suffices to look for poles in the derivative of the numerator of $G(z)$. Using $r = -\sigma_0^2 z e^{\sigma_0^2}$, eqn.~(\ref{eqn:Wr}) and its total derivative with respect to $z$ yields the following equation defining the locations of these poles,
\begin{equation}
e^{W_r}\big(1 + W_r\big) = z e^{z^2}\,,
\end{equation}
which is solved by
\begin{equation}
W_r = W(e^{1+\sigma_0^2} z)-1\,,
\end{equation}
where $W$ is the standard Lambert W function. Next we substite this relation into eqn.~(\ref{eqn:Wr}); zeros in $z$ then define the location of the branch points. Some straightforward algebra yields the maximum and minimum eigenvalue,
\begin{equation}
\lambda_{\pm} = \frac{1}{2} e^{-\frac{1}{2}\sigma_\pm^2}\left(2 + \sigma_\mp^2\right)\,,\quad\text{where}\quad \sigma_\pm^2 = \sigma_0\left(\sigma_0\pm \sqrt{\sigma_0^2 + 4}\right)
\end{equation}

\section{Orthogonal weights are required for stable, universal limiting distributions}

We work at criticality so $\chi = \sigma_w^2 \mu_1 = 1$. This implies that 
\begin{equation}
\begin{split}
\label{eqn:sigmaJJS}
\mu_{JJ^T} & = m_1 = 1\\
\sigma_{JJ^T}^2 &= m_2 - m_1^2 = L \left(\frac{\mu_2}{\mu_1^2} - 1 - s_1\right)\,.
\end{split}
\end{equation}
Observe that Jensen's inequality requires that $\mu_2 \ge \mu_1^2$. If we require that $\sigma_{JJ^T}^2$ approach a constant as $L\to\infty$, we must have that,
\begin{equation}
\label{eqn:s1pos}
s_1\ge 0\,.
\end{equation}
Similarly, writing
\begin{equation}
M_{W^TW}(z) = \sum_{k=1}^{\infty}\frac{\mathfrak{m}_k}{z^k}\,,
\end{equation}
we can relate $\sigma_w$ and $s_1$ to $\mathfrak{m}_1$ and $\mathfrak{m}_2$. Specifically, evaluating the relation,
\begin{equation}
M_{W^TW}^{-1}(z) = \frac{1+z}{z S_{W^TW}(z)}\,,
\end{equation}
at $z = M_{W^TW}(x)$, gives,
\begin{equation}
x = \frac{1 + M_{W^TW}(x)}{M_{W^TW}(x) S_{W^TW}(M_{W^TW}(x))}\,.
\end{equation}
Expanding this equation to second order gives,
\begin{equation}
\begin{split}
\mathfrak{m}_1 &= g^2\\
\mathfrak{m}_2 &= g^2 \mathfrak{m}_1 (1 - s_1)\,.
\end{split}
\end{equation}
Finally we see that,
\begin{equation}
\sigma_{WW^T}^2 = \mathfrak{m}_2 - \mathfrak{m}_1^2 =  -g^4 s_1\,.
\end{equation}
Positivity of variance gives $s_1 \le 0$, which, together with eqn.~(\ref{eqn:s1pos}) implies,
\begin{equation}
s_1 = 0\,.
\end{equation}
Altogether we see that the variance of the distribution of eigenvalues of $WW^T$ must be zero. Since its mean is equal to $\sigma_w^2$, we see that the only valid distribution for the eigenvalues of $WW^T$ is a delta function peaked at $\sigma_w^2$, i.e. the distribution corresponding to the singular values of an orthogonal matrix scaled by $\sigma_w$.

\end{document}